\documentclass[conference]{IEEEtran}
\IEEEoverridecommandlockouts
\usepackage{cite}
\usepackage{amsmath,amssymb,amsfonts}
\usepackage{algorithmic}
\usepackage{graphicx}
\usepackage{textcomp}
\usepackage{booktabs}
\usepackage{multirow}
\usepackage{xcolor}
\def\BibTeX{{\rm B\kern-.05em{\sc i\kern-.025em b}\kern-.08em
    T\kern-.1667em\lower.7ex\hbox{E}\kern-.125emX}}
\begin{document}

\title{CityGen: Structure-Guided City-Style Synthesis for Cross-City Autonomous Driving}

\author{
\IEEEauthorblockN{
Zezhong Qian$^{1,2*}$,
Zhao Yang$^{1,2*}$,
Lu Tan$^{3}$,
Zhihao Yan$^{4}$,
Weiyi Hong$^{2}$,
Haizhuang Liu$^{4}$,
Yawei Jueluo$^{1}$
}

\IEEEauthorblockA{
$^{1}$Jiangsu Cytoderm Intelligent Technology Co., Ltd., China \\
$^{2}$Xi'an Jiaotong University, Xi'an, China \\
$^{3}$Tsinghua University, Beijing, China \\
$^{4}$University of Science and Technology of China, Hefei, China \\
zezhongqian@stu.xjtu.edu.cn, 
lingxi.yz950701@gmail.com, 
gabrielluunn@gmail.com, 
yanzhihao950822@gmail.com, \\
2223315795@stu.xjtu.edu.cn, 
lhzopjang5@outlook.com, 
jueluo.yawei@cytoderm.ai \\
$^*$Equal contribution
}
}

\maketitle


\begin{abstract}
Autonomous driving systems are commonly trained and evaluated within limited geographic regions, which hinders their scalability when deployed in new cities. However, significant domain shifts in appearance, road topology, and traffic patterns often cause severe performance degradation under cross-city deployment. Existing approaches based on domain adaptation, data augmentation, or synthetic data generation typically rely on labeled target data, city-specific annotations, or task-specific designs, limiting their scalability and effectiveness for holistic evaluation. In this paper, we introduce CityTransfer-Bench, a geographically disjoint benchmark for evaluating cross-city generalization across perception, segmentation, and planning, and propose CityGen, a diffusion-based generative framework that performs zero-label city adaptation via HD-map–conditioned synthesis guided by city-level visual prompts. Extensive experiments demonstrate that CityGen consistently improves cross-city robustness across multiple tasks, establishing a scalable and label-efficient foundation for generalizable autonomous driving.
\end{abstract}


\section{Introduction}\label{sec:intro}
Autonomous driving systems are typically trained and evaluated within limited geographic regions, such as specific cities \cite{wang2023exploring, hu2023_uniad, zhou2022cross}. However, this often leads to significant performance degradation when deployed in unseen cities, due to domain shifts in road layouts, visual appearance, traffic patterns, and environmental conditions. Achieving robust \textbf{cross-city generalization} is therefore a critical challenge for scalable and reliable autonomous driving systems. Despite the growing scale and diversity of autonomous driving datasets, systematic evaluation frameworks for assessing and improving model transferability across cities remain limited.

\begin{figure}[t]
\centering
\includegraphics[width=0.47\textwidth]{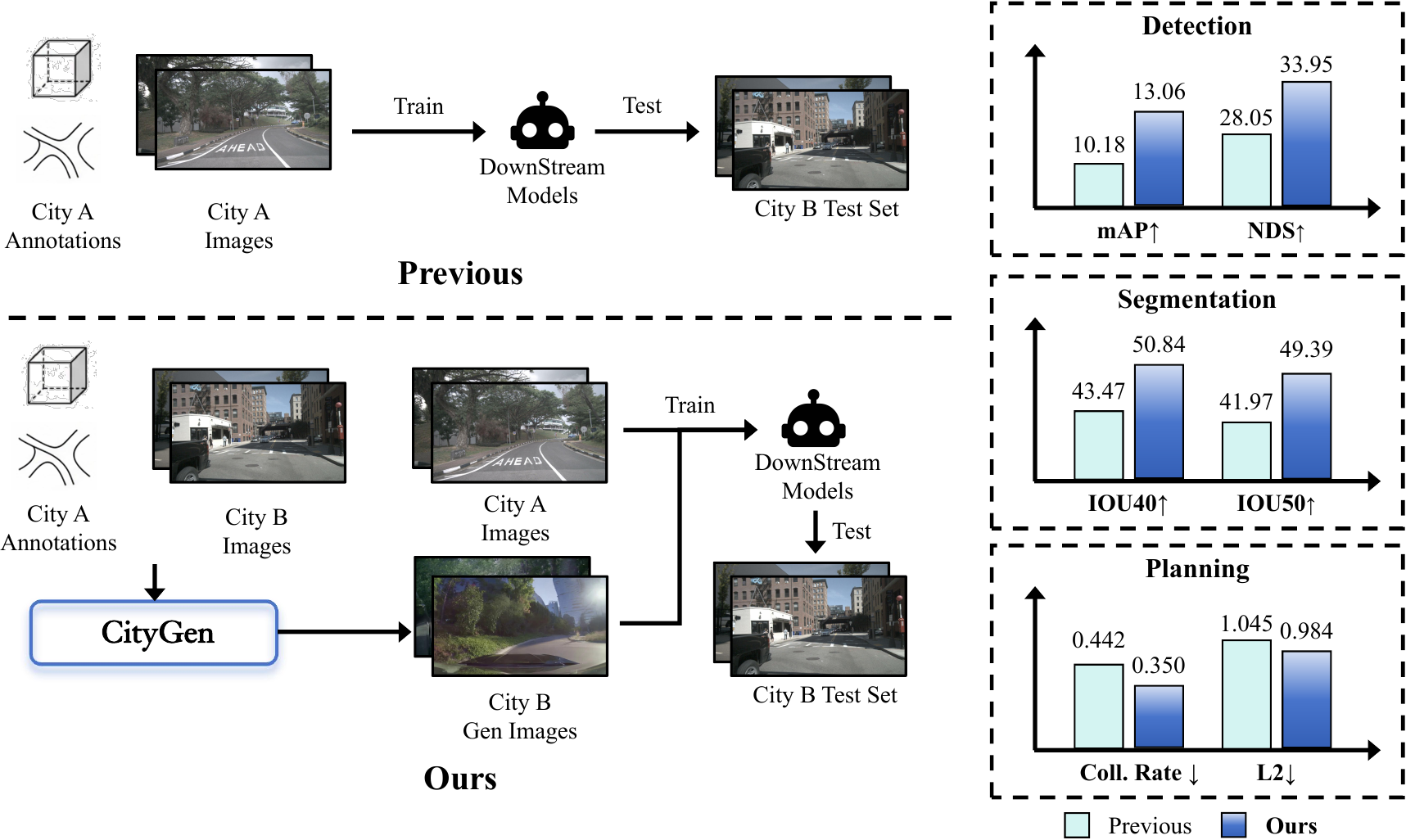}  
\caption{CityGen pipeline and evaluation protocol. Previous methods train models on City A, causing performance degradation on City B. CityGen synthesizes City-B-style images using HD-map-guided generative modeling, improving cross-city generalization in detection, segmentation, and planning.}
\label{fig:tessar}
\vspace{-0.4cm}
\end{figure}

Existing methods have attempted to address cross-city generalization through domain adaptation techniques, data augmentation, and synthetic data generation. However, these methods often rely on either labeled target data or domain-specific annotations, making them less scalable and more expensive. Some methods attempt to align feature distributions between cities through adversarial training or domain-specific fine-tuning \cite{gan2019gated, frank2018cross}, but these techniques still struggle with large-scale, real-world cross-city generalization due to inherent differences in traffic patterns, road structures, and environmental conditions. Additionally, most existing methods focus primarily on single-task performance (e.g., perception or decision-making) without a unified framework that evaluates multi-task generalization across different urban domains.

Designing a benchmark for cross-city generalization is challenging due to urban differences in both low-level visual cues (such as illumination, texture, and architecture) and high-level semantics (such as road topology, lane geometry, and local driving conventions). While existing methods, such as domain adaptation and synthetic data generation \cite{zhu2017unpaired, li2020domain}, attempt to address these issues, they often require extensive annotations or rely on style transfer that may not maintain semantic consistency across cities. Manual annotation for each city is prohibitively expensive, and these methods typically lack scalability across diverse urban environments. Therefore, a scalable and label-efficient framework is essential for evaluating and improving model robustness across cities \cite{RAO202553, Samak_2026, yasarla2025rocarobustcrossdomainendtoend}.

To address these challenges, we introduce \textbf{CityTransfer-Bench}, the first benchmark specifically designed to evaluate cross-city generalization in autonomous driving. Built on the nuScenes dataset, it uses a geographically disjoint split, training on data from three Singapore regions (One-North, Queenstown, and Holland Village) and testing in the Boston Seaport area. The benchmark evaluates three key tasks—perception, segmentation, and decision-making—providing a unified protocol for assessing city-level transferability. 

In conjunction with this benchmark, we propose \textbf{CityGen}, a diffusion-based generative framework for zero-label city adaptation. CityGen uses DiT to synthesize realistic urban scenes conditioned on HD-map layouts and city-specific visual prompts \cite{Peebles2022DiT, nvidia2025cosmostransfer1, nvidia2025cosmosdrivedreams}. This enables semantic-preserving style transfer from Singapore to Boston without additional annotations, making it a plug-and-play solution for improving cross-city robustness.

Our main contributions are as follows:  
(1) We present \textbf{CityTransfer-Bench}, the first benchmark for evaluating city-level generalization across perception, segmentation, and planning tasks in autonomous driving.  
(2) We propose \textbf{CityGen}, a diffusion-based generative framework that performs zero-label city adaptation through HD-map-conditioned synthesis guided by city-level visual prompts.  
(3) We conduct large-scale experiments showing that the data generated by CityGen significantly improves downstream robustness on unseen cities, establishing a scalable foundation for future research in cross-domain autonomous driving.
\vspace{-5pt}
\section{Related work}\label{sec:related}

\textbf{Cross-Domain Generalization in Autonomous Driving.}
Cross-domain generalization is a long-standing challenge in autonomous driving due to
distribution shifts across environments, cities, and weather conditions. Prior work
mainly addresses this issue through domain adaptation and domain generalization
techniques, particularly for perception tasks such as detection and semantic
segmentation \cite{caesar2020nuscenesmultimodaldatasetautonomous,
sun2020scalabilityperceptionautonomousdriving,
cordts2016cityscapesdatasetsemanticurban}. Representative approaches include adversarial
feature alignment, output-space adaptation, and cross-domain data mixing
\cite{hoffman2017cycadacycleconsistentadversarialdomain,
tranheden2020dacsdomainadaptationcrossdomain,
hoyer2022daformerimprovingnetworkarchitectures,
hoyer2022hrdacontextawarehighresolutiondomainadaptive}. Several benchmarks evaluate
robustness under domain shifts such as weather, illumination, and synthetic-to-real transfer
\cite{chung2024acdcautoregressivecoherentmultimodal, Bevandi__2024, 9880320}. However,
most studies rely on labeled target-domain data, focus on single tasks, or use mixed-domain
splits without strict city separation, limiting systematic evaluation of city-level
generalization across autonomous driving tasks.

\textbf{Generative Data Synthesis for Autonomous Driving.}
Recent advances in generative modeling enable realistic driving scene synthesis as an
alternative to conventional data augmentation. Early approaches rely on simulation or
image translation, while recent methods employ diffusion or autoregressive models for
high-fidelity multi-view generation
\cite{wang2023drive, wen2024panacea, gao2025magicdrive-v2,
zhang2025eponaautoregressivediffusionworld}. Many works incorporate structural conditioning
(e.g., layouts, geometry, text prompts) to improve controllability and consistency
\cite{li2023drivingdiffusionlayoutguidedmultiviewdriving,
gao2025magicdrive3dcontrollable3dgeneration}. Generative world models are also explored for
long-horizon prediction and planning \cite{huang2025drivegptscalingautoregressivebehavior}.
However, most approaches emphasize realism or simulation, with limited study on cross-city
generalization. In contrast, our work leverages structure-guided synthesis and evaluates it
under a unified geographically disjoint cross-city benchmark.

\begin{figure}[t]
\centering
\includegraphics[width=0.47\textwidth]{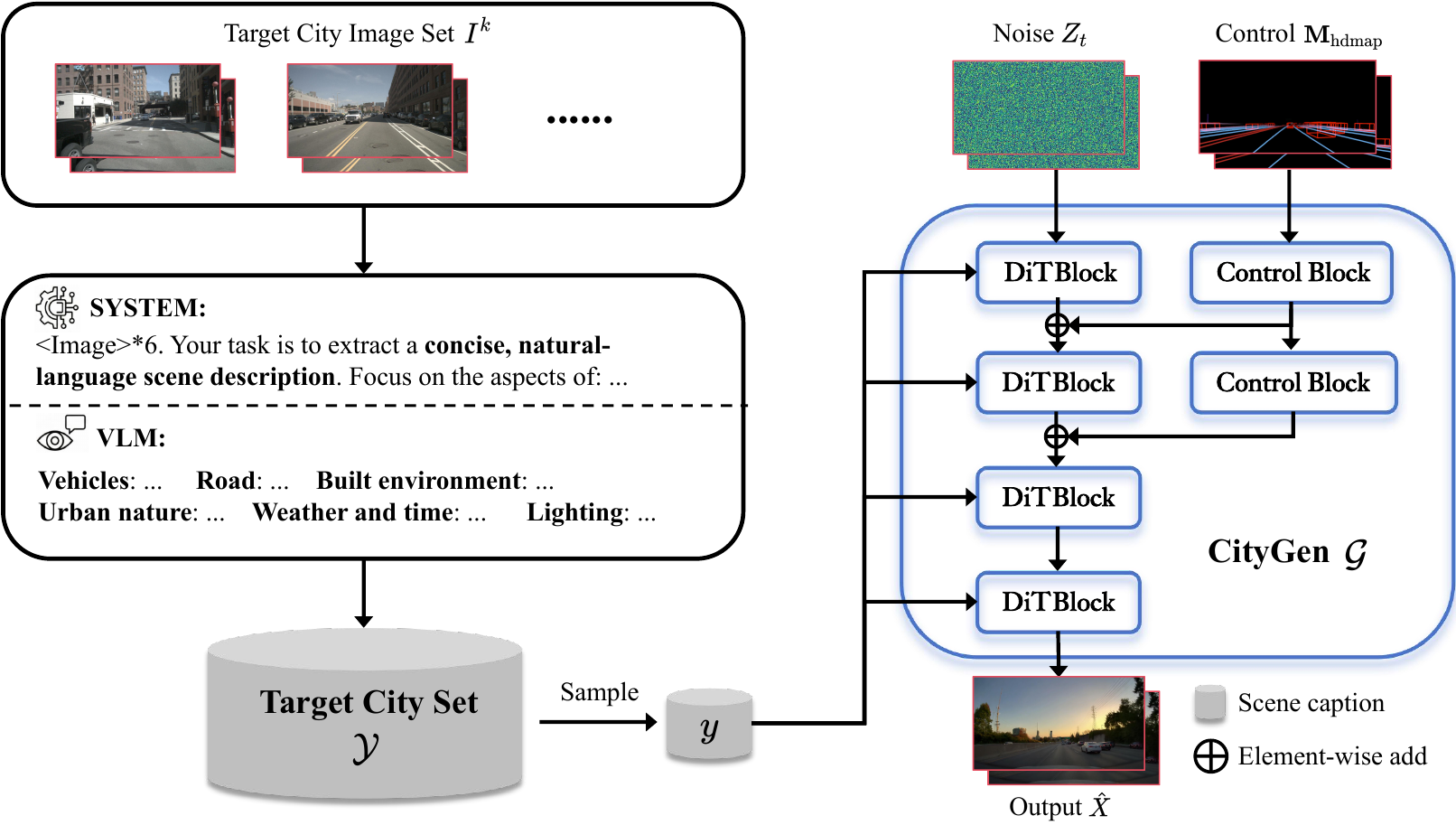}
\caption{CityGen framework. A vision--language model extracts scene captions from target-city multi-view images to form a text-conditioning set. CityGen synthesizes multi-view urban scenes by denoising noise with DiT blocks while injecting structural controls via conditional transfer blocks, producing coherent multi-view images aligned with target-city appearance and layout.}
\label{fig:main}
\vspace{-0.4cm}
\end{figure}

\section{Method}\label{sec:method}

\subsection{Preliminary}\label{sec:preliminary}
{\textbf{Diffusion-based Video Generation with Structural Control.}}  
Given a video $\mathbf{X} = \{x^{t}\}_{t=1}^{T}$, we use a spatiotemporal autoencoder to map it to latent tokens $\mathbf{Z}_{0} \in \mathbb{R}^{T \times C \times H \times W}$. A forward diffusion process progressively adds noise to $\mathbf{Z}_{0}$, producing noisy latents $\mathbf{Z}_{t}$, while a denoising transformer $\epsilon_{\theta}$ is trained to predict the injected noise, conditioned on a signal $\mathbf{c}$:
\begin{equation}
\mathcal{L}_{\mathrm{diff}} = \mathbb{E}_{\mathbf{Z}_{0}, \boldsymbol{\epsilon}, t} \left\| \boldsymbol{\epsilon} - \epsilon_{\theta}(\mathbf{Z}_{t}, t \mid \mathbf{c}) \right\|_2^2.
\end{equation}
The signal $\mathbf{c}$ is injected via cross-attention, guiding the denoising process across both spatial and temporal dimensions.

To enforce structural constraints, we introduce a structural control branch in the diffusion model \cite{zhang2023adding}. For the $\ell$-th transformer block, the backbone features $\mathbf{h}^{\mathrm{main}}_{\ell}$ are modulated by control features $\mathcal{C}_{\ell}(\mathbf{M})$ derived from the structure input $\mathbf{M}$:
\begin{equation}
\mathbf{h}^{\mathrm{final}}_{\ell} = \mathbf{h}^{\mathrm{main}}_{\ell} + \mathcal{C}_{\ell}(\mathbf{M}).
\end{equation}
This structure-preserving modulation enables the generation of outputs that maintain geometric fidelity while retaining the expressiveness of the pretrained diffusion model, yielding the final sampling operator:
\begin{equation}
\hat{\mathbf{X}} = \mathcal{G}(\mathbf{M}, \mathbf{c}).
\end{equation}

\subsection{Structure-Guided Geometric Control via Multi-view HD-map Projection}

As illustrated in Fig.~\ref{fig:main}, CityGen tackles cross-city data synthesis by
explicitly disentangling (i) a city-invariant structural layout and (ii) a city-specific
appearance style. We first introduce the structural control that enforces geometric and
topological consistency across domains. Intuitively, while the visual texture of
buildings, vegetation, and illumination may vary significantly from city to city, the
underlying road topology, lane connectivity, and actor geometry should remain stable
for a given driving sample, thus enabling label preservation.

\textbf{Multi-view structural map construction.}
Given a driving sample with calibrated multi-view cameras, we construct a structural
control tensor $\mathbf{M}_{\mathrm{hdmap}}$ by projecting city-invariant geometric
primitives into each camera plane. The goal of $\mathbf{M}_{\mathrm{hdmap}}$ is to encode
scene layout and object geometry while discarding appearance-dependent factors, thereby
serving as a stable structural anchor for cross-city synthesis.

Specifically, for each camera view $v$, we consider two types of structural elements:
(i) lane-level geometry extracted from the HD-map, including lane boundaries and
centerlines represented as polylines in the world coordinate system, and
(ii) instance-level object geometry represented by 3D bounding boxes of dynamic and
static agents (e.g., vehicles and pedestrians). These primitives are projected into the
image plane using known camera extrinsics and intrinsics, and subsequently rasterized
into dense 2D mask channels. Formally, we define:
\begin{equation}
\mathbf{M}_{\mathrm{hdmap}}=\Big\{\ \mathcal{R}\big(\Pi_v(\mathbf{S})\big)\ \oplus\
\mathcal{R}\big(\Pi_v(\mathbf{B})\big)\ \Big\}_{v=1}^{V},
\end{equation}
where $\mathbf{S}$ denotes lane-related structural elements (e.g., boundaries and
centerlines) represented as polylines, and $\mathbf{B}$ denotes the set of 3D bounding
boxes. $\Pi_v(\cdot)$ is the projection operator for view $v$, mapping 3D geometry from
the world coordinate system to the 2D image plane, and $\mathcal{R}(\cdot)$ rasterizes
the projected primitives into multi-channel binary or soft masks. The operator $\oplus$
denotes channel-wise concatenation. In practice, each view produces a compact stack of
geometry-aware channels (e.g., separate channels for lane boundaries, lane centerlines,
and box silhouettes or edges), and the multi-view stacks are concatenated to form the
final structural control signal $\mathbf{M}_{\mathrm{hdmap}}$.

\textbf{Structure injection into diffusion denoising.}
Following Sec.~\ref{sec:preliminary} and the control pathway illustrated in
Fig.~\ref{fig:main}, the structural map $\mathbf{M}_{\mathrm{hdmap}}$ is processed by a
set of lightweight structural encoders to generate block-wise control features aligned
with the intermediate activations of the diffusion transformer. Concretely, we employ a
collection of encoders $\{\mathcal{C}_\ell\}_{\ell=1}^{L}$, each producing control features
at a specific resolution corresponding to the $\ell$-th denoising block.

At each block, the structural information is injected by feature-wise addition:
\begin{equation}
\mathbf{h}_{\ell}^{\mathrm{final}} = \mathbf{h}_{\ell}^{\mathrm{main}} + \mathcal{C}_{\ell}(\mathbf{M}_{\mathrm{hdmap}}),
\end{equation}
which enforces geometric consistency throughout the entire denoising trajectory rather
than only at the input stage. This design is particularly important for multi-view
synthesis: the projected lane geometry and 3D box structures provide explicit
cross-view correspondences, preventing view-specific drift such as misaligned lane
boundaries or inconsistent object extents under large appearance shifts. As a result,
the generated multi-view urban scenes strictly preserve the spatial layout specified by
$\mathbf{M}_{\mathrm{hdmap}}$, ensuring that existing annotations for perception and
planning tasks remain valid after city-style transfer. In addition, we employ multi-view attention across camera views to explicitly exchange
structural cues between views, further enhancing cross-view geometric consistency.

\subsection{CityGen: Structure-Guided City-Style Synthesis}
With structure rigidly constrained by $\mathbf{M}_{\mathrm{hdmap}}$, CityGen focuses on
synthesizing target-city appearance while preserving the underlying geometry. We model
appearance variations as a learnable ``city-style'' condition that captures domain
factors such as materials, illumination, vegetation density, signage, and architectural
context, which are not encoded by the HD-map but critically affect perception models.

\textbf{Unlabeled city-style library.}
As depicted in Fig.~\ref{fig:main}, we construct a city-style library from unlabeled
target-city videos. The library is defined as:
\begin{equation}
\mathcal{Y} = \left\{ \mathrm{Enc}_{\mathrm{style}}(\mathbf{I}^{k}) \right\}_{k=1}^{K},
\end{equation}
where $\mathbf{I}^{k}$ are sampled frames or short clips from the target domain, and
$\mathrm{Enc}_{\mathrm{style}}$ is a visual--language style encoder implemented with
InternVL. $K$ denotes the number of sampled frames or short clips collected from unlabeled
target-city videos to populate the city-style library. Importantly, the library is built without manual annotations: it aggregates
diverse target-city visual statistics across time (e.g., daytime/night, weather,
seasonal vegetation) and location (e.g., downtown/residential/highway), enabling
controllable and diverse synthesis during training.

\textbf{Style-conditioned diffusion synthesis.}
During generation, we sample a style descriptor $y \sim \mathcal{Y}$ and inject it into
the diffusion transformer as a conditioning signal, jointly with the structural control
$\mathbf{M}_{\mathrm{hdmap}}$. In implementation, the style descriptor modulates the
global appearance direction of the denoising process (e.g., texture, lighting, and color
tone), while the structural control anchors the spatial content (e.g., lane topology,
object positions, and extents) at every denoising block. This separation yields a clean
factorization: structural constraints suppress geometric hallucination, and style
conditioning drives photorealistic domain shift toward the target city. 

The resulting generated sequence is expressed as:
\begin{equation}
\hat{\mathbf{X}} = \mathcal{G}(\mathbf{M}_{\mathrm{hdmap}}, y),
\end{equation}
where $\mathcal{G}$ denotes the diffusion sampling operator. By sampling multiple
$y$ from $\mathcal{Y}$ for the same $\mathbf{M}_{\mathrm{hdmap}}$, we can generate diverse
target-city renderings while keeping the geometry fixed, effectively augmenting the
training distribution along the appearance axis. This is particularly beneficial for
cross-city robustness: downstream models are exposed to a broad spectrum of target-city
visual conditions without requiring any relabeling.

\textbf{Training objective.}
To optimize the generation model, we minimize the diffusion loss during training:
\begin{equation}
\mathcal{L}_{\mathrm{diff}} = \mathbb{E}_{\mathbf{Z}_{0}, \epsilon, t} \left[
\left\| \epsilon_{\theta}(\mathbf{Z}_{t}, t \mid y, \mathbf{M}_{\text{hdmap}}) -
\boldsymbol{\epsilon} \right\|_2^2 \right],
\end{equation}
which encourages accurate denoising under the joint conditioning of city style and
structural layout. Since $\mathbf{M}_{\mathrm{hdmap}}$ is injected into intermediate
blocks, the model is trained to respect geometry not only at reconstruction time but
throughout the entire denoising trajectory, thereby reducing structure violations under
strong appearance shifts. Overall, this formulation enables CityGen to synthesize
multi-view urban scenes that maintain structural integrity while exhibiting distinctive
target-city visual characteristics, improving robustness to appearance shifts and
zero-shot generalization.

\subsection{CityTransfer Benchmark for Cross-City Generalization}

The standard nuScenes protocol trains and evaluates models on scenes from both Singapore
and Boston, which does not explicitly assess cross-city generalization. To address this
limitation, we introduce a geographically disjoint benchmark that trains models
exclusively on all Singapore scenes and evaluates them on all Boston scenes, ensuring
that the target city is entirely unseen during testing.

The benchmark evaluates cross-city robustness across three core components of autonomous
driving: detection, segmentation, and planning. These tasks are instantiated using a
multi-view transformer-based detector, a cross-view representation learning model for
segmentation, and a unified trajectory prediction and decision-making architecture for
planning, enabling a holistic assessment of city-level domain shifts across the
autonomous driving stack.

We define two reference baselines for comparison. $\text{Baseline}_{\text{all}}$ is
trained on the full nuScenes dataset and serves as an approximate upper bound, while
$\text{Baseline}_{\text{s.g.}}$ is trained only on the Singapore subset, measuring
cross-city transferability without any augmentation. For each enhancement method,
synthetic data are used for pretraining, followed by fine-tuning on real Singapore data,
and all models are evaluated on the held-out Boston test set. Together, these settings
constitute the CityTransfer Benchmark, providing a unified protocol for evaluating
city-level generalization in autonomous driving.

\section{Experiments}\label{sec:experiment}
We evaluate CityGen on the CityTransfer Benchmark to assess its impact on cross-city generalization across perception, segmentation, and planning tasks. The downstream models include StreamPetr \cite{wang2023exploring} for multi-view detection, Cross-View Transformer (CVT) \cite{zhou2022cross} for semantic segmentation, and UniAD \cite{hu2023_uniad} for integrated prediction and planning. All experiments are conducted on eight A800 GPUs.

\subsection{Training Details.}
All experiments are conducted on the nuScenes dataset, which contains 28,130 training samples from two cities: Singapore and Boston. Of these, 12,435 samples are from Singapore, and the remaining samples are from Boston. To ensure consistent comparison across downstream models, we adjust the number of training epochs so that the total optimization steps match the original training schedules of each method.

For \textbf{StreamPetr} \cite{wang2023exploring}, we pretrain the model on CityGen-generated data for 20 epochs and fine-tune it on ground-truth data for another 20 epochs. This requires approximately 12 hours for pretraining and 6 hours for fine-tuning on 8$\times$A800 GPUs. \textbf{Cross-View Transformer (CVT)} \cite{zhou2022cross} is pretrained for 17,500 steps on generated samples and fine-tuned for an additional 17,500 steps on ground-truth data, with the full training taking around 12 hours on the same hardware. \textbf{UniAD} \cite{hu2023_uniad} follows its standard Stage~1 training for 40 epochs, followed by 10 epochs of pretraining on CityGen-generated samples and 20 epochs of fine-tuning on annotated data, requiring approximately 20 hours for pretraining and 40 hours for fine-tuning.

\subsection{Main results.}
Across all three tasks, CityGen consistently achieves the best overall performance among
all enhancement methods. For 3D detection, CityGen improves mAP from 10.18 to 13.06 and
NDS from 28.05 to 33.95 compared to $\text{Baseline}_{\text{s.g.}}$, outperforming prior
generative approaches such as Epona and DualDiff+. For BEV segmentation, CityGen yields
particularly large gains, increasing IoU$_{40}$ from 43.47 to 50.84 and IoU$_{50}$ from
41.97 to 49.39, substantially narrowing the gap to the upper-bound
$\text{Baseline}_{\text{all}}$. These improvements can be attributed to two key factors:
first, CityGen enforces strong geometric consistency by conditioning generation on
HD-map–derived structure, which preserves lane topology and object layout across cities;
second, the explicit transfer of target-city visual style exposes downstream models to
the target-domain appearance during training, enabling earlier adaptation to city-level
distribution shifts.

In the planning task, CityGen further reduces the average collision rate to 0.350\% and
the average L2 error to 0.984, outperforming the source-only baseline and achieving
competitive performance relative to the best prior methods. Although planning gains are
more moderate than those observed in perception and segmentation, the same two mechanisms
remain effective: geometry-consistent synthesis ensures physically plausible scene
layouts for trajectory reasoning, while target-city style transfer reduces visual domain
mismatch in the planning inputs, leading to more stable and safer decision-making in
unseen urban environments.

\begin{figure*}[t]
    \centering
    \includegraphics[width=0.8\linewidth]{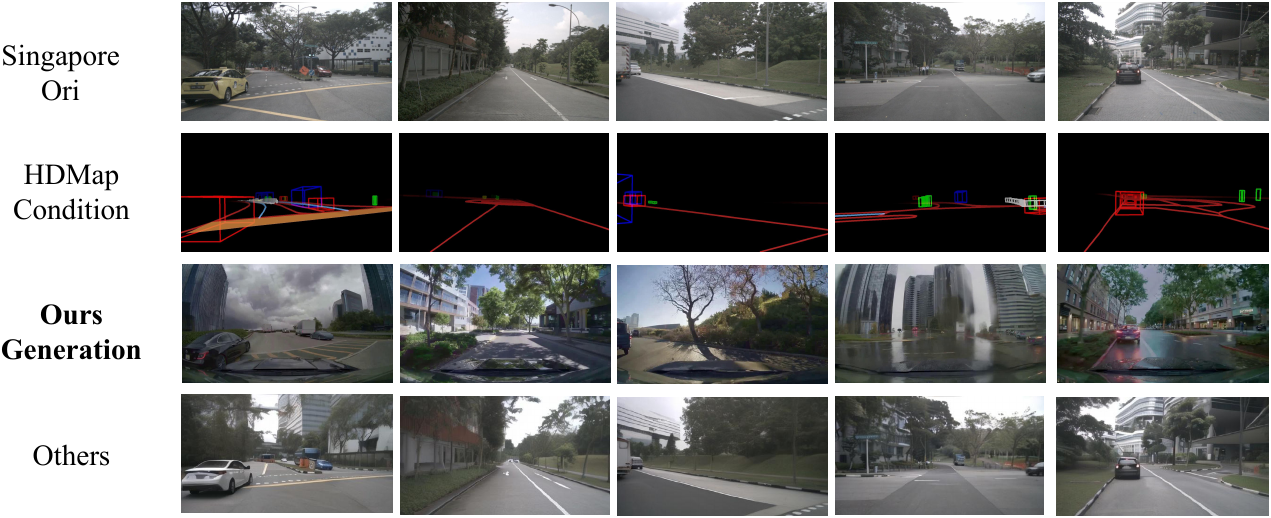}
    \caption{
    Qualitative comparison of CityGen and existing generative methods under cross-city
    settings. CityGen produces visually diverse images with clear target-city style
    characteristics, while other methods exhibit limited deviation from the source-city
    ground-truth style.
    }
    \label{fig:vis}
    \vspace{-0.4cm}
\end{figure*}

\begin{table*}[t]
\centering
\caption{
Comparison of baseline models and enhancement methods across detection,
segmentation and planning tasks on the CityTransfer Benchmark.
The table reports cross-city performance when trained on the training city
and evaluated on the target city. Enhancement models include image-based
augmentations and generative data augmentation baselines. CityGen achieves
gains across all tasks. IoU$_{40}$ and IoU$_{50}$ refer to mean IoU (mIoU) scores for the BEV segmentation task,
computed by averaging per-class IoU over all semantic classes and using IoU thresholds
of 0.40 and 0.50, respectively.
}
\label{tab:main_results}
\begin{tabular}{l|cc|cc|cc}
\toprule
\multirow{2}{*}{Method} 
    & \multicolumn{2}{c|}{\textbf{Detection}} 
    & \multicolumn{2}{c|}{\textbf{Segmentation}} 
    & \multicolumn{2}{c}{\textbf{Planning}} \\
\cmidrule{2-7}
    & $\text{mAP}\uparrow$ & $\text{NDS}\uparrow$ 
    & $\text{IoU}_{40}\uparrow$ & $\text{IoU}_{50}\uparrow$ 
    & $\text{avg.Coll.(\%)}\downarrow$ & $\text{avg.L2}\downarrow$ \\
\midrule

{$\text{Baseline}_{\text{all}}$} 
    & 16.45 & 34.82
    & 76.01 & 75.31
    & 0.373 & 0.871 \\

{$\text{Baseline}_{\text{s.g.}}$} 
    & 10.18 & 28.05
    & 43.47 & 41.97
    & 0.442 & 1.045 \\
\midrule

Enhancement \cite{Song_2024}
    & 11.39 & 32.18 
    & 45.60 & 44.19 
    & 0.353 & 1.034 \\

DriveDreamer \cite{wang2023drive}
    & 9.97 & 30.81
    & 44.48 & 43.04
    & 0.495 & 1.048 \\

MagicDriveV2 \cite{gao2025magicdrive-v2}
    & 10.08 & 27.74
    & 38.32 & 36.60
    & 0.442 & 1.073 \\

Panacea \cite{wen2024panacea}
    & 10.75 & 29.48
    & 39.77 & 37.25
    & 0.555 & \textbf{0.982} \\

DualDiff+ \cite{yang2025dualdiff+}
    & 11.47 & 33.80
    & 43.75 & 42.54
    & 0.360 & 1.039 \\

Epona \cite{zhang2025eponaautoregressivediffusionworld}
    & 12.29 & 32.01
    & 46.85 & 45.49
    & 0.443 & 0.997 \\
\midrule

\textbf{CityGen}
    & \textbf{13.06} & \textbf{33.95}
    & \textbf{50.84} & \textbf{49.39}
    & \textbf{0.350} & 0.984 \\
\bottomrule
\end{tabular}
\vspace{-0.6cm}
\end{table*}

\subsection{Ablation study.}
\noindent{\textbf{Effect of Target-City Style and VLM-based Style Encoding.}}
Table~\ref{tab:ablation_vlm} studies the impact of target-city style alignment and
VLM-based style encoding on cross-city generalization. Replacing target-city captions
with source-city (Singapore) captions leads to consistent performance degradation across
all tasks, reducing detection mAP from 13.06 to 11.24, IoU$_{40}$ from 50.84 to 46.12, and
increasing planning error (avg.L2 from 0.984 to 1.169). This confirms that style priors
must be aligned with the target city, as source-city style tokens bias the diffusion model
toward source-specific appearance statistics, limiting the effectiveness of generated
data for target-domain adaptation. Removing VLM-based captions further degrades
performance (mAP 10.32, IoU$_{40}$ 45.29), indicating that unstructured or low-level style
representations fail to capture semantically meaningful city-level attributes. These
results demonstrate that both target-aligned style conditioning and semantically grounded
VLM-based style encoding are essential for effective city-style transfer and robust
cross-city generalization.

\begin{table}[t]
\centering
\caption{
We examine the effect of replacing target-city style descriptors with those from the training city, as well as removing VLM-based style encoding. Results highlight the necessity of both style alignment and semantic style extraction for cross-city generalization.
}
\label{tab:ablation_vlm}
\resizebox{1.03\linewidth}{!}{
\begin{tabular}{l|cc|cc|cc}
\toprule
\multirow{2}{*}{Method} 
    & \multicolumn{2}{c|}{\textbf{Detection}} 
    & \multicolumn{2}{c|}{\textbf{Segmentation}} 
    & \multicolumn{2}{c}{\textbf{Planning}} \\
\cmidrule{2-7}
    & $\text{mAP}\uparrow$ & $\text{NDS}\uparrow$ 
    & $\text{IoU}_{40}\uparrow$ & $\text{IoU}_{50}\uparrow$ 
    & $\text{avg.Coll(\%).}\downarrow$ & $\text{avg.L2}\downarrow$ \\
\midrule

Singapore captions   & 11.24 & 32.01
    & 46.12 & 44.67
    &  0.436  & 1.169 \\
w/o VLM captions   & 10.32 & 30.18
    & 45.29 & 43.84
    & 0.423 & 1.130 \\
\midrule
\textbf{CityGen}
& \textbf{13.06} & \textbf{33.95}
& \textbf{50.84} & \textbf{49.39}
& \textbf{0.350} & \textbf{0.984} \\
\bottomrule
\end{tabular}
}
\vspace{-0.8cm}
\end{table}

\noindent{\textbf{Effect of Reduced Training Data on CityGen Performance.}}
Table~\ref{tab:data_analysis} evaluates CityGen under limited labeled source data. As the
training set is reduced to 50\% and 20\%, performance degrades as expected due to reduced
coverage and increased overfitting; however, CityGen maintains strong robustness across
all tasks. Even with only 20\% of the training data, CityGen achieves 11.18 mAP in
detection and 47.88 IoU$_{40}$ in segmentation, remaining competitive with full-data
baselines, while the planning error remains controlled (avg.L2 1.147). This robustness stems
from CityGen’s structure-aware synthesis, which augments the training distribution along
the appearance dimension while preserving label consistency through HD-map conditioning.
By exposing downstream models to diverse target-city visual styles under fixed geometry,
CityGen improves label efficiency and reduces reliance on source-specific visual
correlations, enabling more stable cross-city generalization under limited data regimes.

\begin{table}[t]
\centering
\caption{
Training data analysis comparing CityGen with reduced amounts of ground-truth
training data. Even with 20\%--50\% of the original dataset, CityGen maintains strong
performance, demonstrating improved label efficiency and robustness under limited
data regimes.
}
\label{tab:data_analysis}
\resizebox{\linewidth}{!}{
\begin{tabular}{l|cc|cc|cc}
\toprule
\multirow{2}{*}{Method} 
    & \multicolumn{2}{c|}{\textbf{Detection}} 
    & \multicolumn{2}{c|}{\textbf{Segmentation}} 
    & \multicolumn{2}{c}{\textbf{Planning}} \\
\cmidrule{2-7}
    & $\text{mAP}\uparrow$ & $\text{NDS}\uparrow$ 
    & $\text{IoU}_{40}\uparrow$ & $\text{IoU}_{50}\uparrow$ 
    & $\text{avg.Coll(\%).}\downarrow$ & $\text{avg.L2}\downarrow$ \\
\midrule

50\% data    & 12.49 & 33.22
    & 48.78 & 47.14
    & \textbf{0.333} & 1.116 \\
20\% data    & 11.18 & 32.43
    & 47.88 & 45.27
    & 0.367 & 1.147 \\
\midrule
\textbf{CityGen}& \textbf{13.06} & \textbf{33.95}
& \textbf{50.84} & \textbf{49.39}
& 0.350 & \textbf{0.984} \\

\bottomrule
\end{tabular}
}
\vspace{-0.6cm}
\end{table}

\subsection{Visualization.}
Figure~\ref{fig:vis} shows qualitative comparisons between CityGen and representative
generative baselines under the cross-city setting.
Existing methods tend to generate images that remain visually close to the source-city
ground-truth style, resulting in limited stylistic variation and reduced visual diversity,
which provides only marginal benefits for cross-city generalization.
In contrast, CityGen synthesizes diverse urban scenes that capture distinctive
target-city characteristics in terms of architecture, vegetation, illumination, and
overall appearance, while remaining semantically and geometrically consistent with the
HD map.
Such city-style aligned visual diversity substantially enriches the training distribution
and improves the robustness of downstream perception and planning models when transferred
across geographically disjoint cities.

\section{Conclusion}
We introduce \textbf{CityTransfer-Bench}, a benchmark for evaluating cross-city generalization in autonomous driving across perception, segmentation, and planning. Using a geographically disjoint nuScenes split, it reveals significant performance degradation in unseen cities. To address this, we propose \textbf{CityGen}, a diffusion-based framework that synthesizes target-city-style scenes conditioned on HD-map geometry and city-level prompts. Experiments show consistent improvements over existing augmentation and generative baselines, demonstrating that structurally consistent and visually diverse synthesis improves robustness under city-level domain shifts. 

\bibliographystyle{IEEEbib}
\bibliography{icme2026references}

\vspace{12pt}

\end{document}